\documentclass{article}

    \PassOptionsToPackage{numbers, compress}{natbib}
\usepackage{amsmath,amsfonts,bm}

\def\eqref#1{(\ref{#1})}
\def\1{\bm{1}}

\DeclareMathAlphabet{\mathsfit}{\encodingdefault}{\sfdefault}{m}{sl}
\SetMathAlphabet{\mathsfit}{bold}{\encodingdefault}{\sfdefault}{bx}{n}

    \usepackage[final]{neurips_2021}

\usepackage{amsmath}
\usepackage{amsthm}
\usepackage{amssymb}

\usepackage[utf8]{inputenc} % allow utf-8 input
\usepackage[T1]{fontenc}    % use 8-bit T1 fonts
\usepackage{hyperref}       % hyperlinks
\usepackage{url}            % simple URL typesetting
\usepackage{booktabs}       % professional-quality tables
\usepackage{amsfonts}       % blackboard math symbols
\usepackage{nicefrac}       % compact symbols for 1/2, etc.
\usepackage{microtype}      % microtypography
\usepackage[dvipsnames]{xcolor}
\usepackage{environ}
\usepackage{multicol}
\usepackage{multirow}
\usepackage{graphicx}
\usepackage{wrapfig}
\usepackage{tikz}
\usepackage{bbm}
\usepackage{pifont}
\usepackage{colortbl,booktabs}
\PassOptionsToPackage{hyphens}{url}\usepackage{hyperref}
\hypersetup{colorlinks=true}
\usepackage{wrapfig}
\usepackage{algorithmic}
\usepackage{algorithm}
\usepackage{enumitem}

\DeclareMathAlphabet\mathbfcal{OMS}{cmsy}{b}{n}
\newcommand{\Def}[0]{\mathrel{\mathop:}=}

\definecolor{Tianlong_color}{rgb}{0.858, 0.188, 0.478}

\title{Data-Efficient GAN Training Beyond (Just) Augmentations: A Lottery Ticket Perspective}

\author{%
  Tianlong Chen\textsuperscript{1}, Yu Cheng\textsuperscript{2}, Zhe Gan\textsuperscript{2}, Jingjing Liu\textsuperscript{3}, \textbf{Zhangyang Wang\textsuperscript{1}}\\
  {\textsuperscript{1}University of Texas at Austin, \textsuperscript{2}Microsoft Corporation}, \textsuperscript{3}Tsinghua University\\
  \small{\texttt{\{tianlong.chen,atlaswang\}@utexas.edu,\{yu.cheng,zhe.gan\}@microsoft.com}} \\
  \small{\texttt{JJLiu@air.tsinghua.edu.cn}} \\
}

\begin{document}

\maketitle

\begin{abstract}
Training generative adversarial networks (GANs) with limited real image data generally results in deteriorated performance and collapsed models. To conquer this challenge, we are inspired by the latest observation, that one can discover independently trainable and highly sparse subnetworks (a.k.a., lottery tickets) from GANs. Treating this as an inductive prior, we suggest a brand-new angle towards data-efficient GAN training: by first identifying the lottery ticket from the original GAN using the small training set of real images; and then focusing on training that sparse subnetwork by re-using the same set. We find our coordinated framework to offer orthogonal gains to existing real image data augmentation methods, and we additionally present a new feature-level augmentation that can be applied together with them. Comprehensive experiments endorse the effectiveness of our proposed framework, across various GAN architectures (SNGAN, BigGAN, and StyleGAN-V2) and diverse datasets (CIFAR-10, CIFAR-100, Tiny-ImageNet, ImageNet, and multiple few-shot generation datasets). Codes are available at: \url{https://github.com/VITA-Group/Ultra-Data-Efficient-GAN-Training}.

% Training generative adversarial networks (GANs) with limited real image data generally results in deteriorated performance and collapsed models. To conquer this challenge, we are inspired by the latest observation of \citep{Kalibhat2020WinningLT,chen2021gans}, that one can discover independently trainable and highly sparse subnetworks (a.k.a., lottery tickets) from GANs. Treating this as an inductive prior, we suggest a brand-new angle towards data-efficient GAN training: by first identifying the lottery ticket from the original GAN using the small training set of real images; and then focusing on training that sparse subnetwork by re-using the same set. Both steps have lower complexity and are more data-efficient to train. We find our coordinated framework to offer orthogonal gains to existing real image data augmentation methods \cite{karras2020training,zhao2020diffaugment}, and we additionally offer a new feature-level augmentation that can be applied together with them. Comprehensive experiments endorse the effectiveness of our proposed framework, across various GAN architectures (SNGAN, BigGAN, and StyleGAN-V2) and diverse datasets (CIFAR-10, CIFAR-100, Tiny-ImageNet, and ImageNet). Our training framework also displays powerful few-shot generalization ability, i.e., generating high-fidelity images by training from scratch with just 100 real images, without any pre-training. Codes are available at: \url{https://github.com/VITA-Group/Ultra-Data-Efficient-GAN-Training}.
\end{abstract}

\section{Introduction}
\vspace{-1em}
\begin{wrapfigure}{r}{0.45\linewidth}
\centering
\vspace{-1.0em}
\includegraphics[width=1\linewidth]{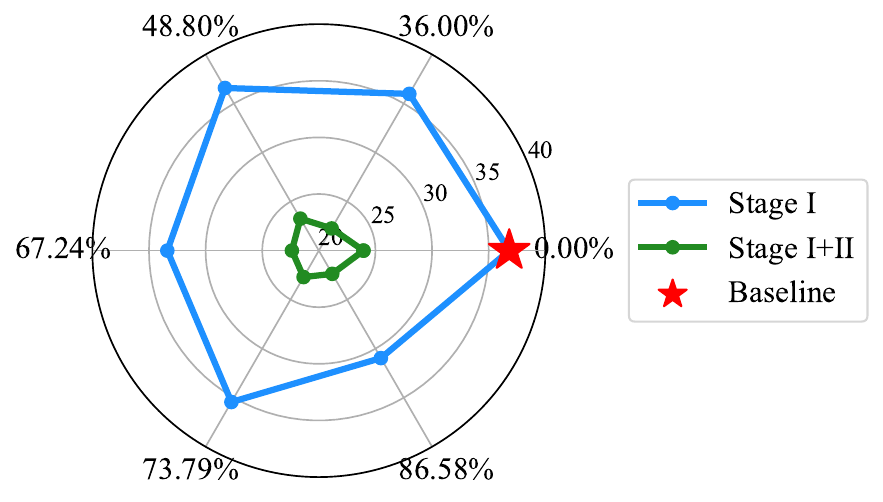}
\vspace{-1.8em}
\caption{\small{FIDs on training BigGAN on $10\%$ training data from CIFAR-100. Smaller distance to the origin indicates smaller FID/better performance. Compared to the vanilla training baseline (\textcolor{red}{\ding{72}}, i.e., dense model or $0\%$ sparsity), our method's Stage I (\textcolor{blue}{$\bullet$}) finds highly sparse lottery tickets from the original BigGAN, with a range of sparsity up to $86.58\%$. Higher sparsity appears to bring better data-efficiency. Stage II further boosts the training of those found sparse subnetworks, by incorporating existing data-level augmentation \cite{zhao2020diffaugment} and our newly proposed feature-level augmentation (\textcolor{OliveGreen}{$\bullet$}).}}
\vspace{-1.5em}
\label{fig:teaser}
% \vspace{-1em}
\end{wrapfigure}

The quantity, diversity, and high quality of natural images available in the general domain have played an essential role in the achieved breakthroughs of Generative Adversarial Networks (GANs)~\cite{brock2018large,goodfellow2014generative,karras2017progressive,karras2019style,karras2020analyzing,Miyato:2018wa,MiyatoK18,kupyn2019deblurgan,gong2019autogan,jiang2021enlightengan} over the past few years.
% Generative adversarial networks (GANs) have achieved breakthroughs  \cite{brock2018large,goodfellow2014generative,karras2017progressive,karras2019style,karras2020analyzing,Miyato:2018wa,MiyatoK18,kupyn2019deblurgan,gong2019autogan,jiang2021enlightengan} over the past few years, blessed by the quantity, diversity, and high quality of natural images available in the general domain. 
However, it could become challenging or even infeasible for specific application domains to collect a sufficiently large-scale dataset, due to various constraints on the imaging expense, subject type, image quality, privacy, copyright status, and more. That prohibits GANs' broader applications in these domains, e.g., for generating synthetic training data \cite{shrivastava2017learning}. Examples of such domains include medical images, images from scientific experiments,  images of rare species, or photos of a specific person or landmark. Eliminating the need of immense datasets for GAN training is highly demanded for those scenarios. Naively training GAN with scarce samples leads to overconfident discriminators that overfit the small training data \cite{arjovsky2017towards,zhang2018pa,karras2020training,zhao2020diffaugment}; it usually ends up with training divergence and drastic performance degradation (evidenced later in Figure~\ref{fig:gan_overfitting}). 

This paper addresses the above issue from a brand new perspective by decomposing the challenging GAN training in limited data regimes into two sequential sub-problems: ($i$) finding independent trainable subnetworks (i.e., lottery tickets in GANs) \cite{chen2021gans,Kalibhat2020WinningLT}; then ($ii$) training the located subnetworks, which we show is more data-efficient by itself, and can further benefit from aggressive augmentations (both the input data and feature levels). 
Either sub-problem becomes much less data-hungry to train, and the two sub-problems re-use the same small training set of real images. Although this paper focuses on tackling the data-efficient training of GANs, such a coordinated framework might potentially be generalized to training other deep models with higher data efficiency too.

Our key enabling technique is to leverage the lottery ticket hypothesis (\textbf{LTH}) \cite{Frankle:2019vz}. LTH shows the feasibility to locate highly sparse subnetworks (called ``\textit{winning tickets}'') that are capable of training in isolation to match or even outperform the performance of original unpruned models. Recently, \cite{Kalibhat2020WinningLT,chen2021gans} revealed the existence of winning ticket in GANs (called ``\textit{GAN tickets}''). However, none of the existing works discuss the influence of training data size on locating and training those tickets. Our work takes one step further, and shows that one can identify the same high-quality GAN tickets even in the data-scarce regime. The found GAN tickets also serve as a sparse structural prior to solve the second sub-problem with less data, while maintaining an unimpaired \textit{trainability} blessed by the LTH assumption \cite{Frankle:2019vz}. Figure~\ref{fig:teaser} (the outer circle's blue dots) evidences that we can identify sparse GAN tickets that achieve superior performance than full GANs in the data-scarce scenarios. 

The new lottery ticket angle complements the existing augmentation techniques \cite{tran2020towards,zhang2019consistency,zhao2020improved}, and we further show that they can be organically combined to boost performance further\footnote{We also tried to add augmentations in the lottery ticket finding stage, but did not observe visible impact.}. When we train the identified lottery ticket, we demonstrate its training can benefit as well from the latest \textit{data-level} augmentation strategies,  ADA~\citep{karras2020training} and DiffAug~\citep{zhao2020diffaugment}. Furthermore, we introduce a novel \textbf{\textit{feature-level} augmentation} that can be applied in parallel to data-level. It injects adversarial perturbations into GANs' intermediate features to implicitly regularize both discriminator and generator. Combining the new feature-level and existing data-level augmentations in training GAN tickets leads to more stabilized training dynamics, and establishes new state-of-the-arts for data-efficient GAN training.

%The next boost comes from extensively exploring augmentation techniques  \cite{tran2020towards,zhang2019consistency,zhao2020improved} in the second sub-problem to train the found GAN ticket. Naively applying data augmentation to GANs leads to the ``leaking issue" \cite{karras2020training} that the generator matches the augmented distribution rather than the true one. \TL{ADA~\citep{karras2020training} and DiffAug~\citep{zhao2020diffaugment} as two pioneer solutions demonstrate appropriate \textit{data-level} augmentation strategies (e.g., adaptive or differentiable) can mitigate the aforementioned problem. From an orthogonal perspective, we introduce the \textit{feature-level} augmentation as an natural remedy, which injects adversarial perturbations into GANs' intermediate features to implicitly regularize both discriminator and generator, avoiding to manipulate the target real data distribution. As shown in Figure~\ref{fig:teaser} (the green cycle), organically combining both data- and feature-level augmentations lead to stabilized training dynamics and improved generation results, which establishes new state of the art for data-efficient GAN training.}

Extensive experiments are conducted on a variety of the latest GAN architectures and datasets, which consistently validate the effectiveness of our proposal. For example, our BigGAN tickets at $36.00\%$ and $67.24\%$ sparsity levels reach an (FID, IS) of $(23.14,52.98)$ and $(70.91,7.03)$, on Tiny-ImageNet $64\times64$ and ImageNet $128\times128$, with $10\%$ and $25\%$ training data, respectively. On CIFAR-10 and CIFAR-100, for SNGAN and BigGAN tickets at $67.24\%\sim86.58\%$ sparsity, our results with only $10\%$ training data can even surpass their dense counterparts. Impressively, our method can generate high-quality images on par with other GAN transfer learning approaches, by training on as few as $100$ real samples and without using any pre-training.

\vspace{-0.5em}
\section{Related Work}
\vspace{-2mm}
\paragraph{GANs and Data-Efficient GAN Training.} GANs \cite{gui2020review} have gained popularity in diverse computer vision scenarios. To stabilize GAN training and improve the visual fidelity and diversity of generated images, extensive studies have been conducted, such as sophisticated network architectures \cite{miyato2018spectral,MiyatoK18,radford2015unsupervised,zhang2019self,chen2020distilling}, improved training recipes \cite{denton2015deep,karras2017progressive,liu2020diverse,zhang2017stackgan}, and more stable objectives \cite{arjovsky2017wasserstein,gulrajani2017improved,mmdgan,mescheder2018training,salimans2016improved,guo2020positive}. \cite{chen2019self,lucic2019high} utilize semi- and self-supervised learning to pursue label efficiency in GAN training. 

%However, the data-efficiency aspect has rarely been explored.

Recently, how to train GANs without sufficient real images in the target domain sparkles new interests. There have been efforts on adapting a pre-trained GAN generator, including BSA \cite{noguchi2019image}, AdaFM \cite{zhao2020leveraging}, Elastic Weight Consolidation \cite{li2020few}, and Few-Shot GAN \cite{robb2020few,antoniou2017data,garcia2017few}. However, those methods assume a
large, related source domain as \textit{pre-training}, based on which they further alleviate target domain data limitation by only tuning small subsets of weights. They are hence in a \textbf{completely different track} from our ``stand-alone'' data-efficient training goal where no pre-training is leveraged. \cite{sinha2020small,devries2020instance} select core-sets of training data to speed up GAN training. A few recent attempts \cite{zhao2020diffaugment,karras2020training} leverage differentiable or adaptive data augmentations to significantly improve GAN training in limited data regimes. Lately, \cite{tseng2021regularizing} investigates a regularization approach, on constraining the distance between the current prediction of the real image and a moving average variable that tracks the historical predictions of the generated image, that complements the data augmentation methods.

\vspace{-1mm}
\paragraph{Lottery Ticket Hypothesis and GAN Tickets.} \cite{Frankle:2019vz} claims the existence of independently trainable sparse subnetworks that can match or even surpass the performance of dense networks. \cite{liu2018rethinking,gale2019state} scale up LTH by rewinding \cite{frankle2019linear,Renda2020Comparing}. Follow-up researches evidence LTH across broad fields, including visual recognition \cite{Frankle:2019vz,liu2018rethinking,grasp,evci2019difficulty,Frankle2020The,savarese2020winning,yin2020the,You2020Drawing,chen2021long,ma2021good,chen2020lottery2,pmlr-v139-zhang21c}, natural language processing \cite{gale2019state,yu2019playing,Renda2020Comparing,chen2020lottery,desai2019evaluating,chen2020earlybert}, graph neural network \cite{chen2021unified}, and reinforcement learning \cite{yu2019playing}. 

Recently, LTH has been extended to GANs by \cite{chen2021gans,Kalibhat2020WinningLT}, who validated the existence of winning tickets in the min-max game beyond minimization. Compared with the aforementioned work, our work is the first to study LTH in the data-scarce regime (for GANs, and in general). Besides finding highly compact yet same capable subnetworks, our work reveals LTH's power in saving training data - an appealing perspective never being examined before.

\vspace{-1mm}
\paragraph{Adversarial Training and Augmentations.} Deep neural networks suffer from severe performance degradation \cite{szegedy2013intriguing,goodfellow2014explaining} when facing adversarial inputs \cite{goodfellow2014explaining, kurakin2016adversarial, madry2017towards}. To address this notorious vulnerability, various defense mechanisms \cite{zhang2019theoretically,schmidt2018adversarially,sun2019towards,nakkiran2019adversarial,stutz2019disentangling,raghunathan2019adversarial,jiang2020robust,wang2020once,hu2020triple,chen2020adversarial} have been proposed. Among others, adversarial training-based approaches achieve superior adversarial robustness \cite{goodfellow2014explaining, kurakin2016adversarial, madry2017towards}, although at the price of sacrificing benign generalization \cite{tsipras2018robustness,zhang2019theoretically,schmidt2018adversarially,sun2019towards,nakkiran2019adversarial,stutz2019disentangling,raghunathan2019adversarial}.

Several recent works investigate enhancing model (benign) generalization ability with adversarial training \cite{xie2020adversarial,zhu2019freelb,wang2019improving,gan2020large,wei2019improved,chen2021alfa}. They adopt adversarially perturbed input images, embeddings, or intermediate features, into model training to ameliorate performance on the clean test sets. Specifically, the damaging effects of adversarial training could be controlled by extra batch normalization \cite{xie2020adversarial} or so. Different from those minimization problems that previous work has focused on, the two-player GAN optimization is more challenging. Generally, the adversarial competition between two players poses impediments to exploit extra adversarial information during training GANs.

\vspace{-0.5em}
\section{Methodology}
\vspace{-0.5em}
% \begin{wrapfigure}{r}{0.40\linewidth}
% \centering
% \vspace{-6mm}
% \includegraphics[width=0.98\linewidth]{Figs/gan_overfitting.pdf}
% \vspace{-4mm}
% \caption{\scriptsize{The performance of SNGAN heavily degrades with a limited amount of training data. Top two figures show that training with $10\%$ of CIFAR-10 data incurs Fr$\acute{\mathrm{e}}$chet Inception Distance (FID) explosion and Inception Score (IS) decrease, and then the model (\textcolor{blue}{\bf{blue curves}}) collapses. Bottom two figures present the discriminator $\mathcal{D}$'s training and validation accuracy of predicting generated images as fake samples.}}
% \label{fig:gan_overfitting}
% % \vspace{-4mm}
% \end{wrapfigure}
\subsection{Revisiting GANs and the Overfitting Challenge}
\vspace{-0.5em}
Generative adversarial networks (GANs) are dedicated to modeling the target distribution with the two-player game formulation of a generator $\mathcal{G}$ and a discriminator $\mathcal{D}$. Specifically, the generator $\mathcal{G}$ takes a random sampled latent vector $\boldsymbol{z}$ (e.g., from a Gaussian distribution)  as input and outputs the fake sample $\mathcal{G}(\boldsymbol{z})$. The discriminator $\mathcal{D}$ aims to distinguish generated fake samples $\mathcal{G}(\boldsymbol{z})$ from real samples $\boldsymbol{x}$. Alternative optimizations for the discriminator's loss $\mathcal{L}_{\mathcal{D}}$ and the generator's loss $\mathcal{L}_{\mathcal{G}}$ are adopted in the standard GAN training, which can be depicted as follows:
\begin{align}
    \begin{array}{ll}
    \mathcal{L}_{\mathcal{D}}&\Def\mathbb{E}_{\boldsymbol{x}\sim p_{\mathrm{data}}(\boldsymbol{x})}[f_{\mathcal{D}}(-\mathcal{D}(\boldsymbol{x}))] + \mathbb{E}_{\boldsymbol{z}\sim p(\boldsymbol{z})}[f_{\mathcal{D}}(\mathcal{D}(\mathcal{G}(\boldsymbol{z})))] \nonumber \\
\mathcal{L}_{\mathcal{G}}&\Def\mathbb{E}_{\boldsymbol{z}\sim p(\boldsymbol{z})}[f_{\mathcal{G}}(-\mathcal{D}(\mathcal{G}(\boldsymbol{z})))],
    \end{array} \label{eq:robust_feature}
\end{align}
where loss functions $f_{\mathcal{D}}(x),f_{\mathcal{G}}(x)$ have multiple choices, e.g., the non-saturating loss \cite{goodfellow2014generative} with $f_{\mathcal{D}}(x)=f_{\mathcal{G}}(x)=\mathrm{log}(1+e^x)$, and the hinge loss \cite{miyato2018spectral} with $f_{\mathcal{D}}(x)=\mathrm{max}(0,1+x)$ and $f_{\mathcal{G}}(x)=x$. $p_{\mathrm{data}}(\boldsymbol{x})$ and $p(\boldsymbol{z})$ represent the data distribution of real samples and latent vectors. $\mathcal{L}_{\mathcal{D}}$ is maximized to update $\mathcal{D}$'s parameters $\phi$ (i.e., $\mathcal{D}(\cdot)\Def\mathcal{D}(\cdot,\phi)$), and $\mathcal{L}_{\mathcal{G}}$ is minimized to update $\mathcal{G}$'s parameters $\theta$ (i.e., $\mathcal{G}(\cdot)=\mathcal{G}(\cdot,\theta)$). 

\begin{multicols}{2}

\begin{algorithm}[H]
% \vspace{10mm}
\caption{Data-Efficient Iterative Magnitude Pruning Procedures}
\label{alg:IMP}
\begin{algorithmic}[1]
    \STATE {\bfseries Input:} Initial two masks $m_g=1^{\|\theta\|_0}$ and $m_d=1^{\|\phi\|_0}$; Initialization weights $\theta_0$ and $\phi_0$
    \STATE {\bfseries Output:} $\{\mathcal{G}(\cdot,\theta_0\odot m_g),\mathcal{D}(\cdot,\phi_0\odot m_d)\}$
    \REPEAT
    \STATE Training $\{\mathcal{G}(\cdot,\theta_0\odot m_g),\mathcal{D}(\cdot,\phi_0\odot m_d)\}$ for $t$ epochs \emph{with limited training data}
    \STATE Pruning $\rho=20\%$ of remaining weights in both $\mathcal{G}$ and $\mathcal{D}$
    \STATE Updating the binary masks $m_g$ and $m_d$ accordingly
    \STATE Rewinding weights of $\mathcal{G}$, $\mathcal{D}$ to $\theta_0$ and $\phi_0$
    \UNTIL{masks reach the desired sparsity level}
\end{algorithmic}
\end{algorithm}

\begin{figure}[H]
\centering
\vspace{-4mm}
\includegraphics[width=0.98\linewidth]{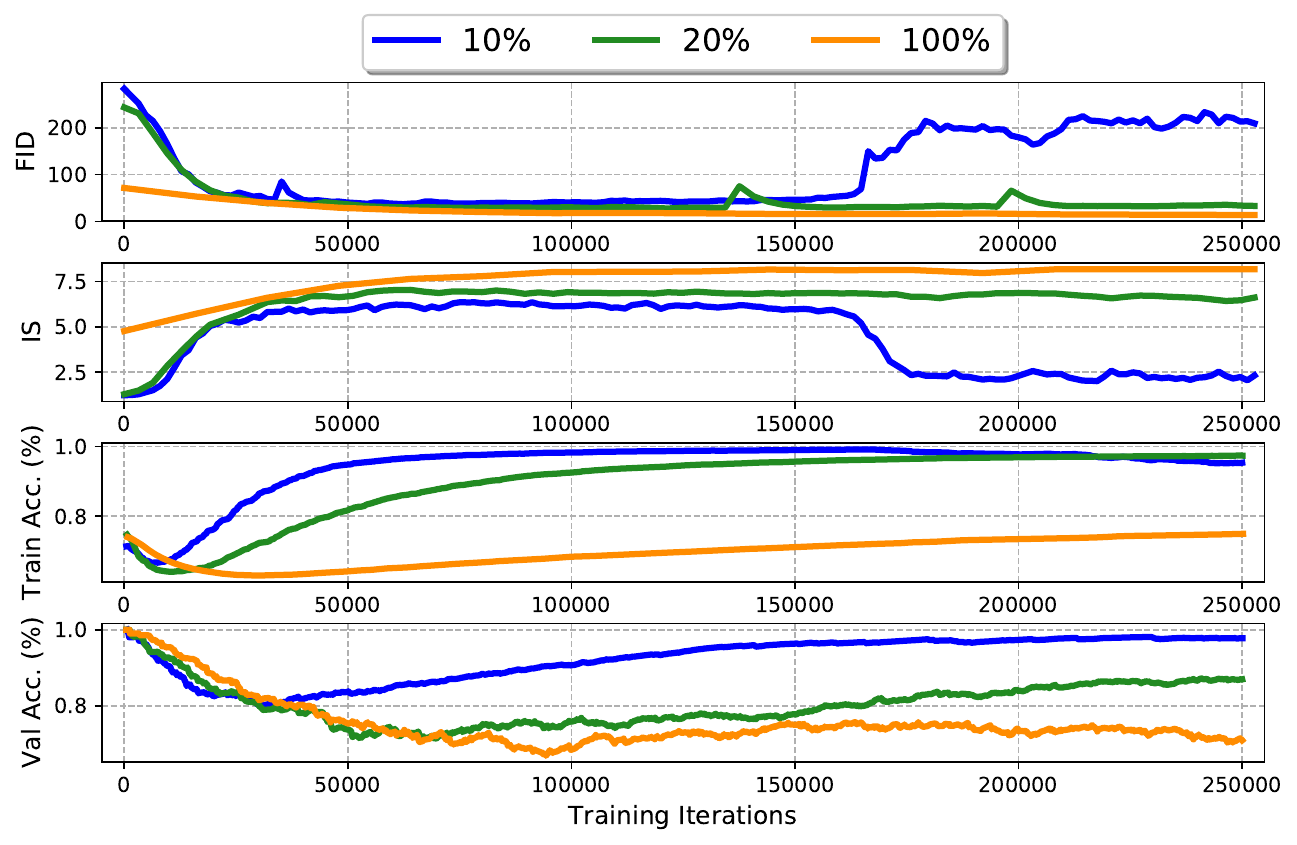}
\vspace{-4mm}
\caption{\small{The performance of SNGAN heavily degrades with limited amount training data. Top two figures show that training with $10\%$ of CIFAR-10 data incurs Fr$\acute{\mathrm{e}}$chet Inception Distance (FID) explosion and Inception Score (IS) drop, with the model (\textcolor{blue}{\bf{blue curves}}) collapsed. Bottom two figures present $\mathcal{D}$'s training and validation accuracies of correctly predicting generated images as fake samples.}}
\label{fig:gan_overfitting}
% \vspace{-4mm}
\end{figure}

\end{multicols}

\paragraph{Training Failures of GANs under Limited Data.} \cite{karras2020training,zhao2020diffaugment} observe that GANs' performance has severely deteriorated when only limited training data is available. The discriminator tends to memorize and heavily \textit{overfit} the small training samples, leaving a gap between the real sample's and the generated sample's distribution. As shown in Figure~\ref{fig:gan_overfitting}, with only $10\%$ CIFAR-10 data available for training SNGAN~\cite{miyato2018spectral}, the training and validation accuracies of the discriminator $\mathcal{D}$ quickly saturate to nearly $100\%$ (ideally close to $50\%$), which indicates $\mathcal{D}$ to become over-confident in distinguishing real and generated samples. It demonstrates that $\mathcal{D}$ simply memorizes the training data, and such overfitting leads to training collapses and deteriorated quality of generated images.

To address this dilemma, we suggest a new data-efficient GAN training workflow, decomposed into two stages: ($i$) \emph{ finding winning tickets in GANs} via Algorithm~\ref{alg:IMP}; then ($ii$) \emph{ training the found GAN tickets}, potentially with both data- and feature-level augmentations, via Algorithm~\ref{alg:AdvAug}. Blessed by LTH, the located GAN ticket shows improved generalization ability, and is further enhanced by augmentations that prevent $\mathcal{D}$ from becoming too confident.

\vspace{-0.5em}
\subsection{Data-Efficient Lottery Ticket Finding from GANs}
\vspace{-0.5em}
In this section, we provide the preliminaries and setups to identifying data-efficient GAN tickets.

\vspace{-0.5em}
\paragraph{Subnetworks and winning tickets.} A subnetwork of GAN is defined as $\{\mathcal{G}(\cdot,\theta\odot m_g),\mathcal{D}(\cdot,\phi\odot m_d)\}$, where $m_g\in\{0,1\}^{\|\theta\|_0}$ and $m_d\in\{0,1\}^{\|\phi\|_0}$ are binary masks for the generator and discriminator respectively, and $\odot$ is the element-wise product. Let $\theta_0$ and $\phi_0$ be the initialization weights of GANs. Following \cite{Frankle:2019vz,chen2021gans}, we define \textit{winning tickets} of GAN as subnetworks $\{\mathcal{G}(\cdot,\theta_0\odot m_g),\mathcal{D}(\cdot,\phi_0\odot m_d)\}$, that reach a matched or better performance compared to unpruned GANs when trained in isolation with similar training iterations. 

\vspace{-0.5em}
\paragraph{Finding data-efficient winning tickets in GANs.} To our best knowledge, we are the first to extend LTH to the limited data regimes. In this challenging scenario, \emph{only a small amount of training data are accessible for the finding and training of GAN tickets}. We use unstructured magnitude pruning \cite{han2015deep}, e.g., Iterative Magnitude Pruning (IMP), to establish the sparse masks $m_g$ and $m_d$.

As shown in Algorithm~\ref{alg:IMP}, we first train the full GAN model for $t$ epochs with limited training samples (e.g., 100-shot), and then perform IMP to globally prune the weights with the lowest magnitude. Zero elements in the obtained masks $m_g$ and $m_d$ index the pruned weights. Before repeating the process again, the weights of the sparse generator $\mathcal{G}(\cdot,\theta\odot m_g)$ and discriminator $\mathcal{D}(\cdot,\phi\odot m_d)$ are rewound to the same initialization $\theta_0$ and $\phi_0$, following the convention \cite{Frankle:2019vz}. The pruning ratio $\rho$ controls the portion of weights removed per round, and we fix $\rho$ = 20\% in all experiments. 

Intuitively, identifying a special sparse mask (without requiring to train its weights well) should be an easier and hence more data-efficient task compared to training the full network weights. That was verified by our observations in experiments too: when the training data volume reduces from $100\%$ to $10\%$ of the full training set, the quality of sparse mask remains to be stable, since it achieves matched performance compared to its dense counterpart in both full and limited data re-training regimes.
% the quality of sparse mask (measured in terms of its achievable re-training performance in full, or limited data regime) remains to be stable when the training data volume reduces from 100\% to 10\% of the full training set.

\vspace{-0.5em}
\subsection{Data-Level and Feature-level Augmentations for Training GAN Tickets}
\vspace{-0.5em}
After locating the GAN ticket at certain sparsity, training it using the vanilla recipe could already attain significantly improved IS and FID compared to the full dense model trained in the same data-limited regime: see Figure~\ref{fig:teaser} outer circle for example. 

Next, we discuss how our proposal can be applied together with augmentation-based approaches, for enhanced training of our found GAN tickets. Our natural choices include to plug-in the two recent state-of-the-art data augmentations, i.e., DiffAug~\citep{zhao2020diffaugment} and ADA~\citep{karras2020training}. We further present a new adversarial \textit{feature-level} augmentation (AdvAug), that can be jointly applied together with data-level augmentations to gain an additional performance boost. 

\begin{figure}[t] 
\vspace{-6mm}
\centering
\includegraphics[width=1\linewidth]{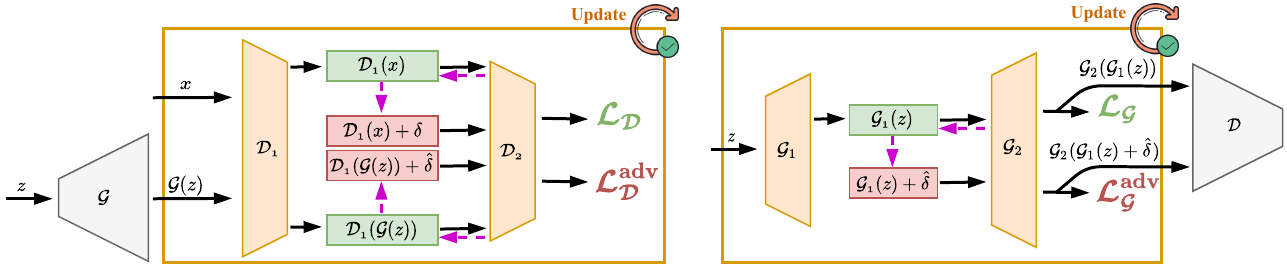}
\vspace{-4mm}
\caption{\small The pipeline of AdvAug for GANs. \textit{Left:} Updating the discriminator $\mathcal{D}$; \textit{Right:} Updating the generator $\mathcal{G}$. Purple arrows denote the path to generate adversarial feature perturbations.}
\vspace{-5mm}
\label{fig:AdvAug_gan}
\end{figure}

\paragraph{Revisiting adversarial training.} Let $(\boldsymbol{x},\boldsymbol{y})$ denote the input image and its label. $f(\vartheta,\boldsymbol{x},\boldsymbol{y})$ is the loss function parameterized by $\vartheta$. Adversarial training \cite{madry2017towards} can be formulated as follows:
\begin{align}
\begin{array}{ll}
\displaystyle \min_{\vartheta}\mathbb E_{(\boldsymbol{x}, \boldsymbol{y})} \displaystyle
\left[\max_{\|\delta\|_{\mathrm{p}}\le\epsilon}f(\vartheta,\boldsymbol{x}+\delta,\boldsymbol{y})\right],
\end{array}
\end{align}
where $\delta$ is crafted adversarial perturbation constrained within the $\ell_{\mathrm{p}}$ norm ball that is centered at $x$ with a radius $\epsilon$. $\delta$ can be reliably generated by multi-step projected gradient descent (PGD) \cite{madry2017towards}. Different from the above standard adversarial training, which adds perturbation on the image pixel space, AdvAug injects adversarial perturbations to intermediate feature embeddings of both $\mathcal{G}$ and $\mathcal{D}$. A similar feature augmentation idea was proven to be helpful in NLP \cite{zhu2019freelb}, and computer vision \cite{chen2021alfa}, showing effectiveness to regularize the smoothness of the training landscape and enhance the trained model's generalization. The AdvAug scheme is illustrated in Figure \ref{fig:AdvAug_gan}, and can be mathematically depicted as follows ($\lambda_1$ is a controlling hyperparameter): 
\begin{align}
\min_{\theta}\mathcal{L}_\mathcal{G}+\lambda_1\cdot\mathcal{L}^{\mathrm{adv}}_{\mathcal{G}} \,\, s.t. \,\, \mathcal{L}^{\mathrm{adv}}_{\mathcal{G}}\Def\max_{\|\hat\delta\|_{\infty}\le\epsilon}\mathbb{E}_{\boldsymbol{z}\sim p(\boldsymbol{z})}[f_{\mathcal{G}}(-\mathcal{D}(\mathcal{G}_2(\mathcal{G}_1(\boldsymbol{z})+\hat\delta)))].
%&\min_{\theta}\mathcal{L}_\mathcal{G}+\lambda_1\cdot\mathcal{L}^{\mathrm{adv}}_{\mathcal{G}}, 
\end{align}
We choose $\lambda_1=1$ in all experiments for simplicity. $\mathcal{G}=\mathcal{G}_2\circ\mathcal{G}_1$ denotes the generator, and between the $\mathcal{G}_1$ and $\mathcal{G}_2$ parts we inject AdvAug. Adversarial perturbations $\hat\delta$ generated by PGD~\cite{madry2017towards}, are applied to the intermediate feature space $\mathcal{G}_1(\boldsymbol{z})$. The details of our feature-level augmentation on the discriminator $\mathcal{D}$ is included in Appendix~\ref{sec:more_methods}. The full algorithm of training GAN with both \textbf{data- and feature-level augmentations} is summarized in Algorithm~\ref{alg:AdvAug}.

\begin{algorithm}[!ht]
\caption{Training (Sparse) GAN with Data- and Feature-level Augmentations}
\label{alg:AdvAug}
\begin{algorithmic}
    \STATE {\bfseries Input:} GAN $\{\mathcal{G}(\cdot,\theta_0),\mathcal{D}(\cdot,\phi_0)\}$; Inputs $\boldsymbol{x}$ and $\boldsymbol{z}$
    \STATE {\bfseries Output:} Trained GAN $\{\mathcal{G}(\cdot,\theta_\mathrm{T}),\mathcal{D}(\cdot,\phi_\mathrm{T})\}$
    \FOR{$t=1$ {\bfseries to} $\mathrm{T}$}
        \STATE \textcolor{gray}{\# Training discriminator with data and feature augmentations} 
        \STATE Augment input with DiffAug~\citep{zhao2020diffaugment} or ADA~\citep{karras2020training}
        \STATE Feed $\boldsymbol{x}$ and $\mathcal{G}(\boldsymbol{z})$ to $\mathcal{D}$
        \STATE Generate adversarial augmented features in $\mathcal{D}$ ( Equation.~\ref{eq:adv_d})
        \STATE Update the discriminator $\mathcal{D}(\cdot,\phi_t)$ (Equation.~\ref{eq:max_adv_d})
        \STATE \textcolor{gray}{\# Training generator with data and feature augmentations}
        \STATE Sample and augment $\boldsymbol{z}$ with DiffAug~\citep{zhao2020diffaugment} or ADA~\citep{karras2020training}
        \STATE Feed $\boldsymbol{z}$ to $\mathcal{G}$. Generate adversarial augmented features in $\mathcal{G}$ ( Equation.~\ref{eq:adv_g})
        \STATE Update the discriminator $\mathcal{G}(\cdot,\theta_t)$ (Equation.~\ref{eq:min_adv_g})
    \ENDFOR
\end{algorithmic}
\end{algorithm}

Note that AdvAug only affects the generated images through $\mathcal{G}$ intermediate features, and the classifier learning through $\mathcal{D}$ features. It hence avoids to directly manipulate the real data distribution. One bonus of doing so is that it is potentially better at alleviating the distribution leaking issue \cite{karras2020training}, i.e., GANs learn to mimic and generate the augmented distribution rather than the real one.

\vspace{-0.5em}
\section{Experiments} \label{sec:exp_details}
\vspace{-0.5em}
In this section, we conduct comprehensive experiments on Tiny-ImageNet \cite{wu2017tiny}, ImageNet \cite{deng2009imagenet}, CIFAR-10 \cite{krizhevsky2009learning}, and CIFAR-100 based on the unconditional SNGAN \cite{miyato2018spectral} and StyleGAN-V2 \cite{karras2020analyzing}, as well as the class-conditional BigGAN \cite{brock2018large}. We adopt the common evaluation metrics, including Fr$\acute{\mathrm{e}}$chet Inception Distance (FID) \cite{heusel2017gans} and Inception Score (IS) \cite{salimans2016improved}. Note that a smaller FID $(\downarrow)$ and a larger IS $(\uparrow)$ indicate better performing GAN models. Furthermore, we evaluate our proposed method on few-shot generation both with and without pre-training in Section~\ref{sec:fewshot}. Extensive ablation studies analyze effectiveness of each component in Section~\ref{sec:ablation}.

\vspace{-0.5em}
\paragraph{Implementation and Baseline Details.} We follow the popular StudioGAN codebase \cite{kang2020ContraGAN}, which contains high-quality re-implementation of BigGAN and SNGAN on ImageNet and CIFAR. For example, our implemented BigGAN baseline performs much better, i.e., FID: $26.44$ (ours) v.s. $39.78$ (reported) on CIFAR-10, and FID: $36.58$ (ours)  v.s. $66.71$ (reported) on CIFAR-100, than the recent reported baselines in \cite{zhao2020diffaugment}, under $10\%$ training data regimes. For detailed configuration, BigGAN takes learning rates of $\{4,2,2\}\times10^{-4}$ for $\mathcal{G}$, of $\{1,5,2\}\times10^{-4}$ for $\mathcal{D}$, batch sizes of $\{256,256,64\}$, $1\times10^{5}$ training iterations, and $\{1,2,5\}$ $\mathcal{D}$ steps per $\mathcal{G}$ step on \{Tiny-ImageNet, ImageNet, CIFAR\} datasets. SNGAN uses learning rates of $2\times10^{-4}$ for $\mathcal{G}$ and $\mathcal{D}$, batch sizes of 64, $5\times10^{4}$ training iterations, and five $\mathcal{D}$ steps per $\mathcal{G}$ step on CIFAR. For StyleGAN-V2 experiments, we use its popular PyTorch implementation\footnote{\href{https://github.com/lucidrains/StyleGAN-V2-pytorch}{https://github.com/lucidrains/StyleGAN-V2-pytorch}. Note that the PyTorch version remains with a small performance gap compared to the TensorFlow implementation in \cite{zhao2020diffaugment}.}, and keep the default configuration in \cite{zhao2020diffaugment} including image resolution ($256\times256$), learning rates for $\mathcal{D}$/$\mathcal{G}$ ($2\times10^{-4}$), batch size (5), and training iterations ($1\times10^{5}$). 

Note that, same as the setting in \cite{zhao2020diffaugment}, training iterations will be doubled when training GANs with DiffAug. We use implementations in the StudioGAN codebase for DiffAug~\citep{zhao2020diffaugment}, and the official implementation\footnote{\href{https://github.com/NVlabs/StyleGAN-V2-ada-pytorch}{https://github.com/NVlabs/StyleGAN-V2-ada-pytorch}. The official Pytorch implementation in~\citep{karras2020training}.} for ADA~\citep{karras2020training}. AdvAug with PGD-1 and step size $0.01/0.001$ is applied on CIFAR/(Tiny-)ImageNet datasets, which are tuned by a grid search in Section~\ref{sec:ablation}. All GANs are trained with $8$ pieces of NVIDIA V100 32GB.

\vspace{-0.5em}
\subsection{On the Effectiveness of Training with Winning Ticket and AdvAug} \label{sec:exp}
\vspace{-0.5em}
\begin{table}[!ht]
\caption{\small \textbf{Tiny-ImageNet} $64\times64$ performance without the truncation trick \cite{brock2018large}. FID and IS are measured using 10K samples; the official validation set is utilized as the reference distribution. BigGANs at $0.00\%$ (full unpruned models), $36.00\%,67.24\%$ sparsity are found and trained with $100\%,20\%,10\%$ data, respectively.}
\vspace{-2mm}
\label{tab:tinyimagenet}
\centering
\small
\resizebox{1\textwidth}{!}{
\begin{tabular}{l|cccccc}
\toprule
\multirow{2}{*}{Methods} & \multicolumn{2}{c}{100\% training data (full set)} & \multicolumn{2}{c}{20\% training data} & \multicolumn{2}{c}{10\% training data}\\ \cmidrule(lr){2-3} \cmidrule(lr){4-5} \cmidrule(lr){6-7}
& FID ($\downarrow$) & IS ($\uparrow$) & FID ($\downarrow$) & IS ($\uparrow$) & FID ($\downarrow$) & IS ($\uparrow$)\\ \midrule
BigGAN (0.00\%) &  21.54 $\pm$ 0.03  & 18.33 $\pm$ 0.15 & 59.77 $\pm$ 0.05 & 7.81 $\pm$ 0.20 & 84.53 $\pm$ 0.08 & 5.45 $\pm$ 0.23  \\
+ AdvAug & 21.07 $\pm$ 0.03 & 18.92 $\pm$ 0.09 & 58.55 $\pm$ 0.05 & 8.46 $\pm$ 0.19 & 81.72 $\pm$ 0.05 & 6.32 $\pm$ 0.18 \\ \midrule
BigGAN (36.00\%) &  20.54 $\pm$ 0.05 & 18.42 $\pm$ 0.20 & 59.56 $\pm$ 0.04 & 7.98 $\pm$ 0.20 & 75.76 $\pm$ 0.08 & 6.49 $\pm$ 0.21\\
+ AdvAug & \textbf{20.02 $\pm$ 0.04} & \textbf{19.15 $\pm$ 0.18} & 58.24 $\pm$ 0.05 & 8.55 $\pm$ 0.20 & 71.47 $\pm$ 0.07 & 6.86 $\pm$ 0.20  \\ \midrule
BigGAN (67.24\%) &  26.37 $\pm$ 0.03 & 16.38 $\pm$ 0.15 & 59.02 $\pm$ 0.03 & 8.17 $\pm$ 0.18 & 73.23 $\pm$ 0.05 & 6.68 $\pm$ 0.15\\
+ AdvAug & 25.59 $\pm$ 0.03 & 17.62 $\pm$ 0.16 & \textbf{57.60 $\pm$ 0.04} & \textbf{8.94 $\pm$ 0.18} & \textbf{70.91 $\pm$ 0.05} & \textbf{7.03 $\pm$ 0.16} \\
\bottomrule
\end{tabular}
}
\vspace{-2mm}
\end{table}

%\paragraph{Tiny-ImageNet and ImageNet}

We adopt the top-performing model BigGAN \cite{brock2018large}, and report experiments on both Tiny-ImageNet at $64\times64$ resolution and ImageNet at $128\times128$ resolution. We evaluate our proposal on Tiny-ImageNet with $10\%,20\%,100\%$ data available, and ImageNet with $25\%$ data available, with results summarized in Tables~\ref{tab:tinyimagenet} and~\ref{tab:imagenet}, respectively. All our results are averaged over three independent evaluation runs (same hereinafter), and the best performance of each column are highlighted. 

The following observations can be drawn: \underline{\textit{First}}, sparse BigGAN tickets can achieve consistently improved performance over the full model ($0.00\%$).
%which demonstrates the broad existence of data-efficient BigGAN winning tickets with different portions of training data of Tiny-ImageNet and ImageNet. 
Especially, with only $10\%$ training data available, BigGAN tickets at $67.24\%$ sparsity obtain massive gains of $11.30$ FID and $1.58$ IS on Tiny-ImageNet. \underline{\textit{Second}}, feature-level augmentation (AdvAug) consistently improves all training cases, from dense to sparse, and from full data to limited data. In particular, larger sparsity (e.g.,  $67.24\%$) with less training data available (e.g., $10\%$) tend to benefit more from applying AdvAug, which is aligned with our design principle. \underline{\textit{Third}}, while the full data regime ($100\%$ data) does not necessarily prefer the highest sparsity (moderate sparsity still benefits), the limited data regimes ($10\%$ or $20\%$ data) see monotonically increasing gains as the ticket sparsity goes higher. That is understandable since the former may need more model capacity to absorb full training data, while the latter case hinges on sparsity to avoid overfitting their limited training data.

\begin{wraptable}{r}{0.48\linewidth}
\vspace{-2.5em}
\caption{\small \textbf{ImageNet} $128\times128$ performance without the truncation trick \cite{brock2018large}. FID and IS are measured using 50K samples; the validation set is utilized as the reference distribution. BigGANs at $0.00\%$ and $36.00\%$ sparsity levels are adopted, and only $25\%$ training data are available in all training stages.}
\label{tab:imagenet}
\centering
\small
\resizebox{0.47\textwidth}{!}{
\begin{tabular}{l|cc}
\toprule
\multirow{2}{*}{Methods} & \multicolumn{2}{c}{25\% training data} \\ \cmidrule(lr){2-3}
& FID ($\downarrow$) & IS ($\uparrow$) \\ \midrule
BigGAN ($0.00\%$) & 25.37 $\pm$ 0.07 &  46.50 $\pm$ 0.40  \\
+ AdvAug & 23.95 $\pm$ 0.06 & 47.95 $\pm$ 0.32\\ \midrule
BigGAN ($36.00\%$) &  24.03 $\pm$ 0.08  & 50.07 $\pm$ 0.51  \\
+ AdvAug & \textbf{23.14} $\pm$ \textbf{0.07} & \textbf{52.98} $\pm$ \textbf{0.47} \\ 
\bottomrule
\end{tabular}
}
\vspace{-3em}
\end{wraptable}

\begin{figure}[t] 
\centering
\includegraphics[width=1\linewidth]{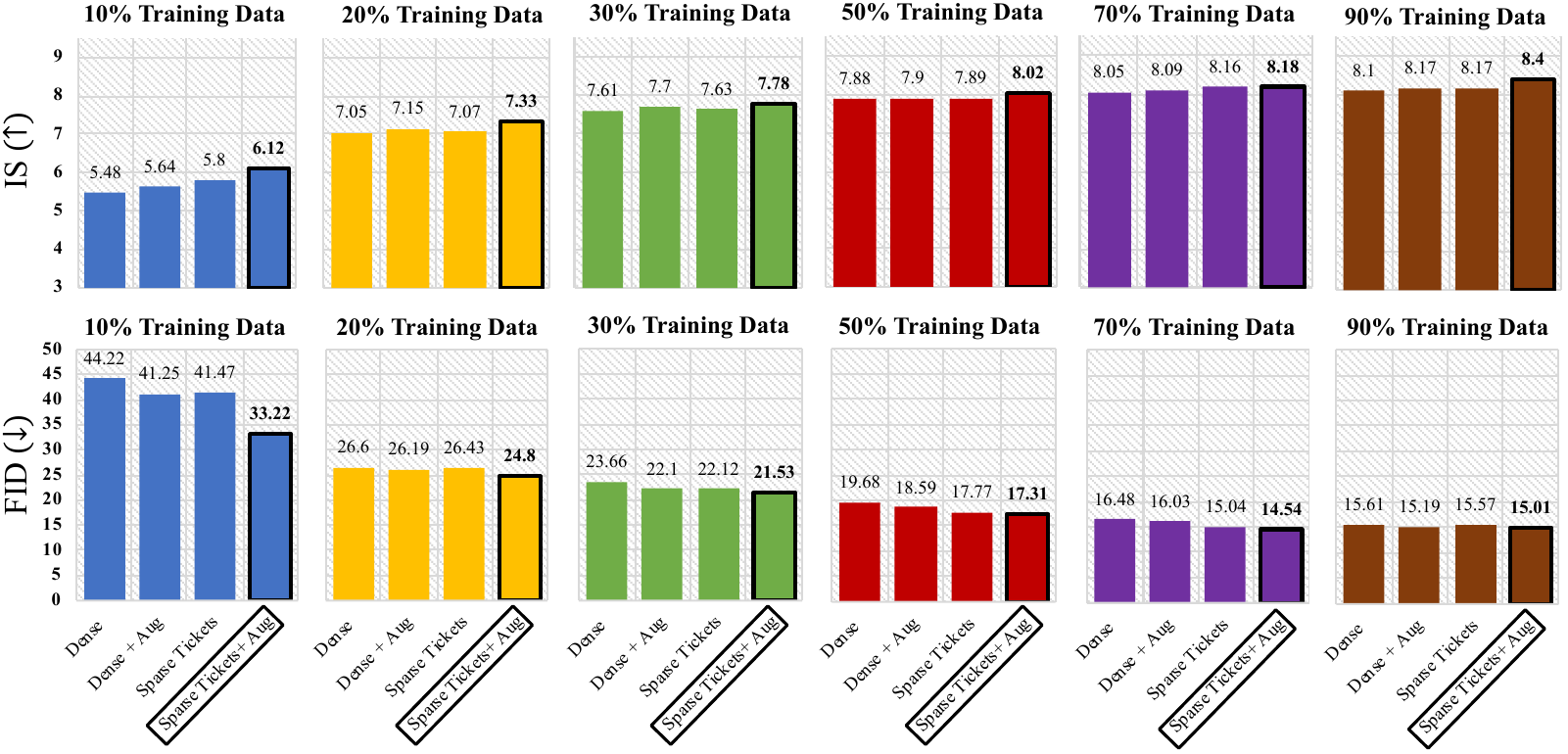}
\vspace{-6mm}
\caption{{\small IS ($\uparrow$) and FID ($\downarrow$) results of SNGAN with $10\%,20\%,30\%,50\%,70\%,90\%$ training data of \textbf{CIFAR-10}. Four settings are evaluated:($i$) Dense (unpruned SNGAN), ($ii$) Dense+Aug (we only apply AdvAug here), ($iii$) Sparse Tickets (pruned SNGAN), ($iv$) Sparse Tickets+Aug, where the top performing variants are highlighted with \textbf{black boxes}. SNGAN tickets with $20\%,36\%,36\%,67\%,49\%,36\%$ sparsity levels are adopted accordingly. 
%We apply the feature-level augmentation, i.e., AdvAug, to enhance data-efficient GAN training. 
IS and FID are measured using 10K samples; the validation set is utilized as the reference.\vspace{-0.5em}}}
\label{fig:SNGAN}
\end{figure}

Besides, we report another group of experiments of training SNGAN on CIFAR-10, using from $10\%$ to $90\%$ training data. The results are summarized in Figure~\ref{fig:SNGAN}. The conclusions we can draw are highly consistent with the above BigGAN case: (1) at the same data availability (from 10\% to even 90\%), training a sparse ticket is always preferred over training the dense model; (2) AdvAug is also consistently helpful in all cases; (3) for both sparsity and AdvAug, they can contribute to larger gains when training data gets smaller.

%such intriguing sparse BigGANs can be further enhanced by training with the feature-level augmentation (AdvAug) as shown in both Table~\ref{tab:tinyimagenet} and~\ref{tab:imagenet}, and sparser GAN tickets (e.g., at $67.24\%$ sparsity) with less training data available (e.g., $10\%$ training data) tend to benefit more from applying AdvAug. \underline{\textit{Third}}, our two-step decomposition pipeline is much more effective on limited data regime compared to sufficient training data setup, i.e., $10.81$ FID reduction on $10\%$ training data v.s. $1.05$ FID reduction on $100\%$ training data. A possible explanation is that training GAN with fewer data incurs severer overfitting (Figure~\ref{fig:gan_overfitting}), which potentially benefits more from the implicit regularization of our proposed sparse structure prior, i.e., winning tickets.

\begin{table}[t]
\caption{\small \textbf{CIFAR-10 and CIFAR-100} results. FID ($\downarrow$) are measured using 10K samples; the validation set is utilized as the reference distribution. Full dense models and sparse winning tickets of BigGAN are reported with $100\%,20\%,10\%$, respectively. Specifically, BigGAN tickets with $67.24\%$ and $86.58\%$ sparsity levels are reported for the $100\%$, $20\%$ and $10\%$ training data regimes. Performance reported is averaged over three independent evaluation runs; all standard deviation are less than $1\%$.}
\vspace{-2mm}
\label{tab:cifar}
\centering
\small
%\resizebox{1\textwidth}{!}{
\begin{tabular}{l|cccccc}
\toprule
\multirow{2}{*}{Methods} & \multicolumn{3}{c}{CIFAR-10} & \multicolumn{3}{c}{CIFAR-100} \\ \cmidrule(lr){2-4} \cmidrule(lr){5-7}
& 100\% data & 20\% data & 10\% data & 100\% data & 20\% data & 10\% data \\ \midrule
\textbf{Dense} BigGAN  & 8.57 & 17.38 & 26.44 & 11.83 & 22.13 & 36.58 \\
+ DiffAug \cite{zhao2020diffaugment} & 8.09 & 13.04 & 17.40 & 10.60 & 18.32 & 25.69 \\
+ DiffAug + AdvAug & \textbf{7.70} & 12.19 & 14.40 & \textbf{8.96} & 17.94 & 23.94\\ \midrule
\textbf{Sparse} BigGAN Tickets & 8.26 & 16.03 & 25.41 & 11.73 & 21.05 & 30.96 \\
+ DiffAug \cite{zhao2020diffaugment} & 8.19 & 12.83 & 16.74 & 10.73 & 17.43 & 23.80\\
+ DiffAug + AdvAug & 8.15 & \textbf{12.02} & \textbf{14.38} & 10.14 & \textbf{17.19} & \textbf{22.37} \\
\bottomrule
\end{tabular}
%}
\vspace{-2mm}
\end{table}

% \vspace{-0.5em}
\subsection{Incorporating Our Proposal with Latest Data Augmentations and Regularization}
\vspace{-0.1em}

\begin{wraptable}{r}{0.48\linewidth}
\vspace{-5mm}
\caption{\small \textbf{CIFAR-100} results with only $10\%$ training data available. FID ($\downarrow$) of three evaluation runs are measured using 10K samples; the validation set is utilized as the reference distribution. Sparse StyleGAN-V2 tickets at $48.80\%$ sparsity are adopted.}
\label{tab:ada_styleganv2}
\centering
\small
\resizebox{0.47\textwidth}{!}{
\begin{tabular}{l|c}
\toprule
\multirow{1}{*}{Methods} & \multicolumn{1}{c}{10\% training data} \\ \midrule
\textbf{Dense} StyleGAN-V2 & 13.59 $\pm$ 0.06   \\
+ DiffAug~\citep{zhao2020diffaugment} & 12.90 $\pm$ 0.04  \\ 
+ ADA~\citep{karras2020training} & 12.87 $\pm$ 0.03   \\ 
+ ADA + $R_{\mathrm{LC}}$~\citep{tseng2021regularizing} & 13.01 $\pm$ 0.02  \\ \midrule
\textbf{Sparse} StyleGAN-V2 Tickets & 13.05 $\pm$ 0.07  \\
+ DiffAug~\citep{zhao2020diffaugment} & 12.53 $\pm$ 0.03  \\
+ ADA~\citep{karras2020training} & 12.20 $\pm$ 0.03  \\ 
+ ADA + $R_{\mathrm{LC}}$~\citep{tseng2021regularizing} & 12.48 $\pm$ 0.04  \\
+ ADA~\citep{karras2020training} + AdvAug & \textbf{12.11} $\pm$ \textbf{0.05}  \\
% + ADA + $R_{\mathrm{LC}}$~\citep{tseng2021regularizing} + AdvAug & \textcolor{red}{TBA}  \\
% + ADA~\citep{karras2020training} + AdvAug + $R_{\mathrm{LC}}$  & \textcolor{red}{OPTIONAL}  \\
\bottomrule
\end{tabular}
}
\vspace{-6mm}
\end{wraptable}

\paragraph{Combining DiffAug.} We first incorporate DiffAug \cite{zhao2020diffaugment}, as a representative of the latest data-level augmentation, into our proposal and show the complementary gains. We conduct experiments on the class-conditional BigGAN and unconditional SNGAN models with CIFAR-10 and CIFAR-100. For BigGAN, we utilize $100\%,20\%,10\%$ data to locate GAN tickets; then we train them with data-level DiffAug, or with both DiffAug and feature-level AdvAug, as shown in Table~\ref{tab:cifar}. Consistent observations can be drawn: \underline{\textit{First}}, similar to our previous observations on AdvAug, DiffAug also shows to contribute more when the training data becomes more limited; \underline{\textit{Second}}, combining DiffAug and AdvAug improves over either alone, and leads to the best results across all cases.

\paragraph{Combining Other Data Augmentations and Regularization.} We then extend our combination study to other recent data augmentation and regularization approaches, e.g., ADA \citep{karras2020training} and $R_{\mathrm{LC}}$ \cite{tseng2021regularizing}. Experiments are conducted on CIFAR-100 with StyleGAN-V2 backbone, and results are collected in Table~\ref{tab:ada_styleganv2}. We observe that plugging in either ADA~\citep{karras2020training} or DiffAug~\cite{zhao2020diffaugment} into our framework could improve sparse GAN winning tickets, and the gain is also enlarged when ADA is combined with AdvAug. Regard to $R_{\mathrm{LC}}$\footnote{\citep{tseng2021regularizing} advocates the best-performing configuration is ADA+$R_{\mathrm{LC}}$, with FID $13.01$ on $10\%$ data of CIFAR-100.}, it is less effective combined with other augmentations.

% \textcolor{red}{Same observations are made with regard to $R_{\mathrm{LC}}$: it alone can improve sparse ticket and can boost further when combined with data/feature augmentations.} 

Taking above together, it has been clearly shown that our proposal is orthogonal to those existing efforts and is of independent merit. Moreover, combing them would lead to more powerful pipelines for data-efficient GAN training.

%\TL{We further evaluate and apply our two-step pipeline with other recent data augmentation approaches~\citep{karras2020training,tseng2021regularizing} . Experiments are conducted on CIFAR-100 with StyleGAN-V2 backbone, and results are collected in Table~\ref{tab:ada_styleganv2}. We observe that pluging either ADA~\citep{karras2020training} or DiffAug~\cite{zhao2020diffaugment} into our framework enjoy the performance improvement from sparse GAN winning tickets. Moreover, additional gains are obtained when incorporating with feature-level augmentation, AdvAug. It demonstrates that our coordinated framework can be effective with diverse augmentation mechanisms, leading to enhanced generalization ability for data-efficient GAN training.}

\subsection{Few-Shot Generation} \label{sec:fewshot}

It is laborious, and sometimes impossible to collect a large-scale dataset for certain images of interest. To tackle the few-shot image generation problem, \cite{wang2018transferring} utilizes pre-training from external large-scale datasets and performs fine-tuning under limited data scenarios; \cite{mo2020freeze}, \cite{noguchi2019image} and \cite{wang2020minegan} partially fine-tune the GANs with part of the GAN model being frozen. 

We compare these transfer learning approaches\footnote{Implementations are from the codebase of \cite{mo2020freeze}.} with our data-efficient training scheme. \textbf{Differently from them}, ours is training from scratch and is free of any pre-training, while all transfer learning methods start from a pre-trained StyleGAN-V2 model on the FFHQ face dataset \cite{karras2019style}.

\begin{figure}[H]
\centering
% \vspace{-1em}
\includegraphics[width=1\linewidth]{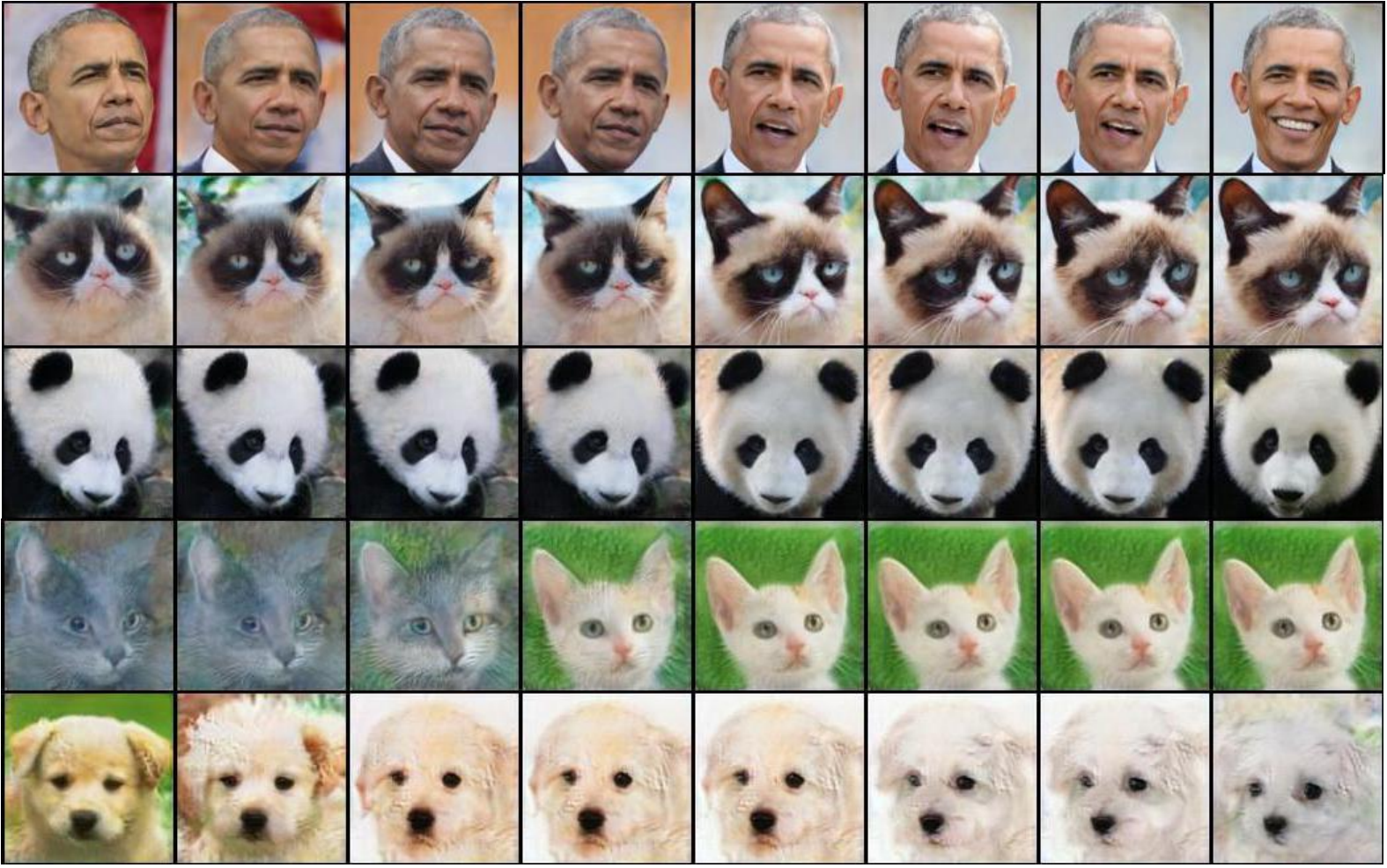}
\vspace{-3mm}
\caption{\small Style interpolation visualizations of StyleGAN-V2 tickets ($48.80\%$) with AdvAug only on 100-shot Obama, Grumpy Cat, Panda, and AnimalFace datasets, respectively.}
\vspace{-1em}
\label{fig:vis_nopre}
\end{figure}

\begin{figure}[H] 
\centering
\includegraphics[width=1\linewidth]{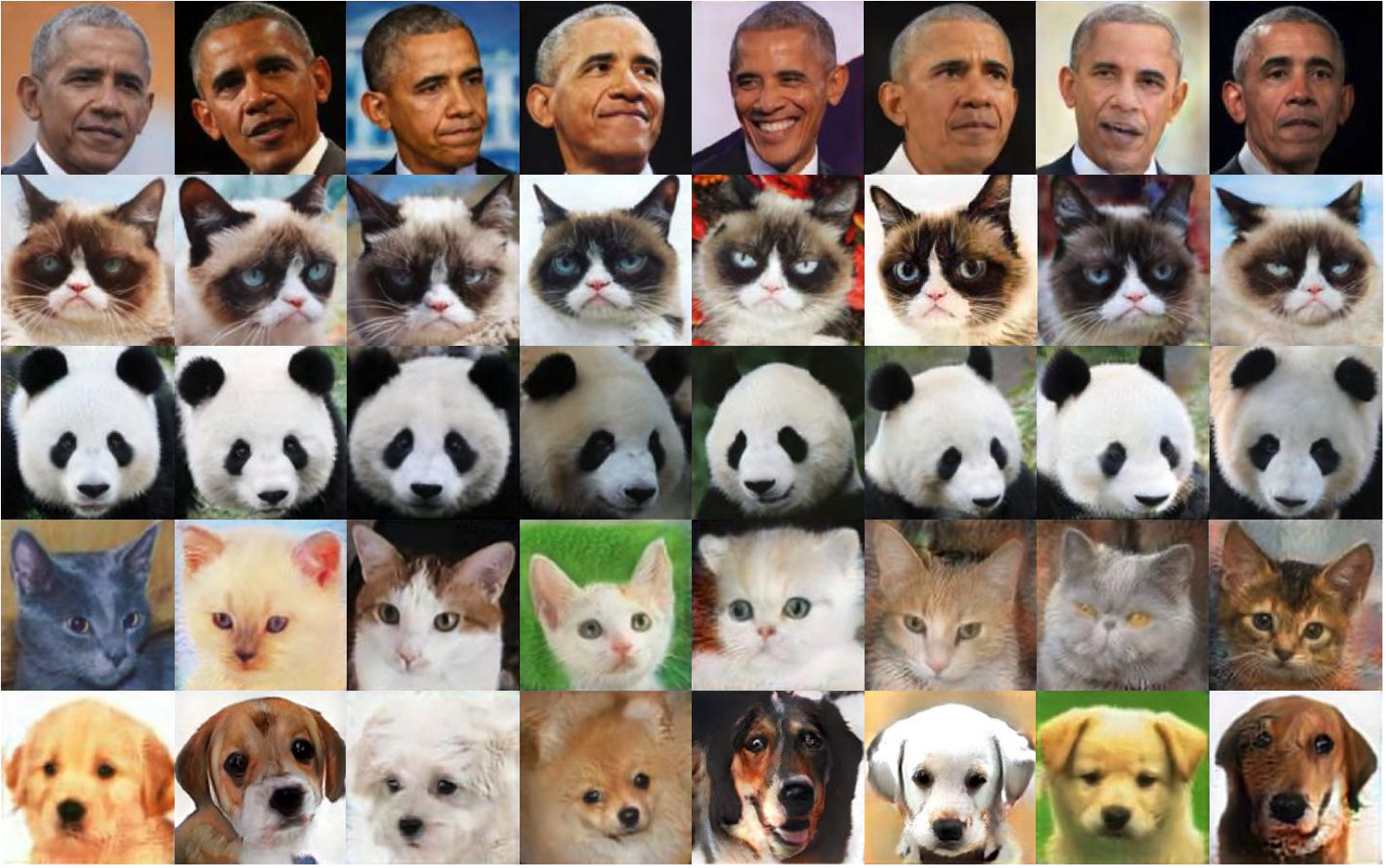}
\vspace{-3mm}
\caption{\small Few-shot generalization results of StyleGAN-V2 tickets ($48.80\%$) with AdvAug only on 100-shot Obama, Grumpy Cat, Panda, and AnimalFace datasets, respectively.}
\vspace{-1em}
\label{fig:100shot_generation}
\end{figure}

Our comparison experiments are conducted using StyleGAN-V2 on the AnimalFace \cite{si2011learning} dataset ($160$ cats and $389$ dogs), and the 100-shot Obama, Grumpy Cat, and Panda datasets provided by \cite{zhao2020diffaugment}. As shown in Table~\ref{tab:fewshot}, our method finds data-efficient GAN tickets at $48.80\%$ sparsity levels, that can be trained with only $100$ training samples from scratch (\textit{without any pre-training}) and show competitive performance to other transfer learning algorithms. Visualizations of style space interpolation and few-shot generation are provided in Figure~\ref{fig:vis_nopre} and~\ref{fig:100shot_generation}.

\begin{table}[!ht]
\caption{\small \textbf{Few-shot generation.} Following the setting in \cite{zhao2020diffaugment}, we calculate the FID with 5K samples and the training dataset is adopted as the reference distribution. All transfer learning methods have their pre-trainings from FFHQ \cite{karras2019style}. StyleGAN-V2 tickets at $48.80\%$ sparsity level are found and used in our method.}
\vspace{-2mm}
\label{tab:fewshot}
\centering
\small
%\resizebox{1\textwidth}{!}{
\begin{tabular}{lcccccc} 
\toprule
\multirow{2}{*}{Methods} & \multirow{2}{*}{Pre-training?} & \multicolumn{3}{c}{100-shot by \cite{zhao2020diffaugment}} & \multicolumn{2}{c}{AnimalFace} \\ \cmidrule(lr){3-5} \cmidrule(lr){6-7}
&  & Obama & Grumpy Cat & Panda & Cat & Dog \\ \midrule
Scale/shift \cite{noguchi2019image} & Yes & 50.72 & 34.20 & 21.38 & 54.83 & 83.04 \\
MineGAN \cite{wang2020minegan} & Yes & 50.63 & 35.54 & 14.84 & 54.45 & 93.03 \\ 
TransferGAN \cite{wang2018transferring} & Yes & 48.73 & 34.06 & 23.20 & 52.61 & 82.38 \\
FreezeD \cite{mo2020freeze} & Yes & 41.87 & 31.22 & 17.95 & 47.70 & 70.46\\ \midrule
StyleGAN-V2 ($0.00\%$) & No & 89.18 & 61.97 & 90.96 & 95.75 & 164.54\\
+ DiffAug + AdvAug & No & 54.11 & 35.46 & 15.94 & 54.02 & 72.47\\ \midrule
StyleGAN-V2 Tickets ($48.80\%$) & No & 73.92 & 56.81 & 82.45 & 85.92 & 153.90\\
+ DiffAug + AdvAug & No & \textbf{52.86} & \textbf{31.02} & \textbf{14.75} & \textbf{47.40} & \textbf{68.28}\\ 
\bottomrule
\end{tabular}
%}
\vspace{-1.5mm}
\end{table}

\begin{multicols}{2}

\end{multicols}

\subsection{Ablation and Analysis} \label{sec:ablation}
\begin{figure}[t] 
\centering
\includegraphics[width=1\linewidth]{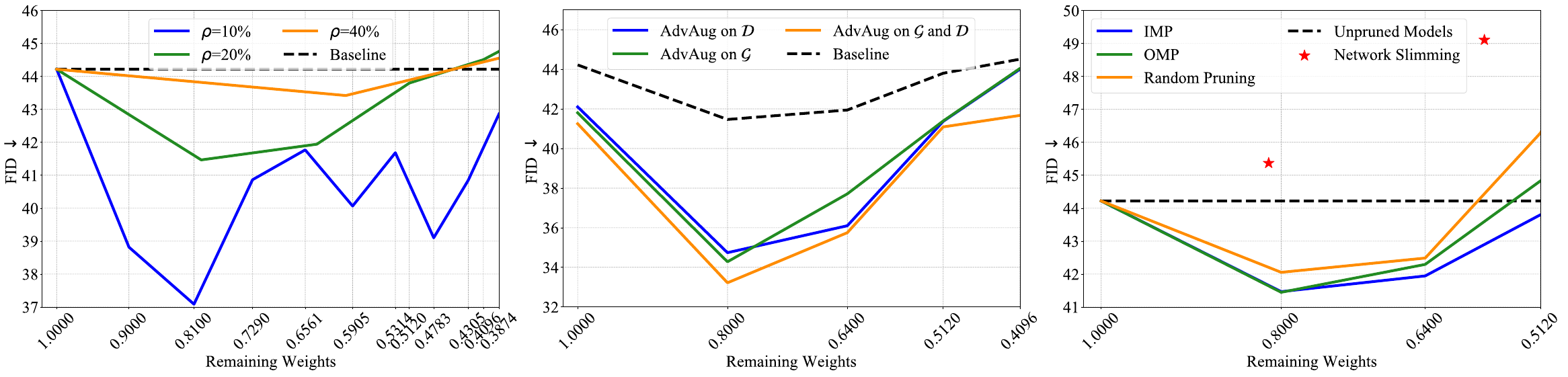}
\vspace{-6mm}
\caption{{\small Performance of training GAN with $10\%$ data of CIFAR-10. Remaining weight indicates the sparsity levels of identified GAN tickets. \textit{Left}: FID of the data-efficient GAN tickets found by IMP with pruning ratios $\rho=10\%,20\%,40\%$.} {\small \textit{Middle}: FID of the data-efficient GAN tickets trained with different settings of AdvAug, including baseline without AdvAug, AdvAug on $\mathcal{D}$ or $\mathcal{G}$ only, and AdvAug on both $\mathcal{G}$ and $\mathcal{D}$.} {\small \textit{Right}: FID of trained SNGAN tickets found by IMP, Random Pruning, OMP, and Network Slimming \cite{liu2017learning}.}}
\label{fig:aba}
\vspace{-3mm}
\end{figure}

\vspace{-0.5em}
\paragraph{Pruning Ratio $\rho$ in the Ticket Finding.} To understand the effect of the pruning ratio in IMP to the quality of data-efficient GAN tickets, we experiment on SNGAN with $10\%$ data of CIFAR-10 and $\rho=10\%,20\%,40\%$ as the pruning ratio. As shown in Figure~\ref{fig:aba} (\textit{Left}), all three IMP settings find data-efficient winning tickets in GAN; IMP with a lower pruning ratio tends to identify higher-quality GAN tickets in terms of FID, while it usually costs much more to reach the same level of sparsity as higher pruning ratios do.

\vspace{-0.5em}
\paragraph{Augment $\mathcal{G}$ or $\mathcal{D}$: Either or Both.} We apply AdvAug on $\mathcal{G}$ or $\mathcal{D}$ only, and AdvAug on both $\mathcal{G}$ and $\mathcal{D}$. Only $10\%$ data of CIFAR-10 are available for finding GAN tickets and training with AdvAug. Results are summarized in Figure~\ref{fig:aba} (\textit{Middle}). Either employing AdvAug on $\mathcal{D}$ or $\mathcal{G}$ consistently obtains significant performance improvements (i.e., largely reducing the FID) over all sparsity levels, and augmenting both $\mathcal{D}$ and $\mathcal{G}$ further enhance the found data-efficient GAN tickets. Our results also show that sparser GAN tickets can benefit more from AdvAug, such as subnetworks with $20\%,36\%,83.22\%$ sparsity.

\begin{figure}[t] 
\centering
\includegraphics[width=1\linewidth]{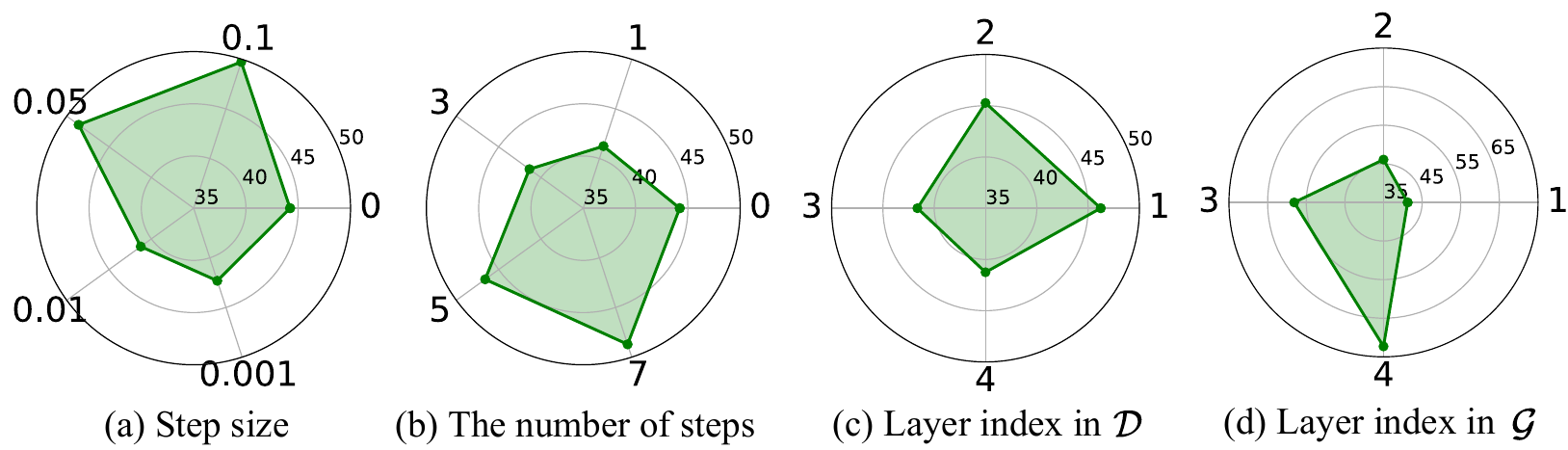}
\vspace{-6mm}
\caption{\small Ablation study on the location and strength of introducing AdvAug to data-efficient GAN training. The step size and the number of steps roughly indicate the strength of generated adversarial perturbations, e.g., a smaller step size or fewer steps for PGD means less aggressive perturbations \cite{madry2017towards}. FID ($\downarrow$) is reported.}
\label{fig:ablation}
\vspace{-1em}
\end{figure}

\vspace{-0.5em}
\paragraph{Strength and Locations of Injecting AdvAug.} To better interpret the influence of the strength and layer locations of injected adversarial feature perturbations, we comprehensively examine SNGAN on $10\%$ training data of CIFAR-10 across different step sizes, the number of PGD steps, and locations (i.e., where to apply AdvAug). When studying one of the factors, we fix the other factors with the best setup. From Figure~\ref{fig:ablation}, several observations can be drawn:
\vspace{-1mm}
\begin{itemize}
    \item Figure~\ref{fig:ablation} (a) and (b) show that adopting AdvAug with step size 0.01/0.001 and PGD-1/3 assists the data-efficient GAN training, while AdvAug with step size 0.05/0.1 and PGD-5/7 perform worse than the baseline without AdvAug (i.e., the setting with zero step size or zero step PGD in Figure~\ref{fig:ablation}). It reveals that overly strong AdvAug can hurt performance. \vspace{-1mm}
    \item As shown in Figure~\ref{fig:ablation} (c) and (d), augmenting the last layer of the discriminator $\mathcal{D}$ and the first layer (i.e., the closest layer to the latent input vector) of the generator $\mathcal{G}$ appears to be the best configuration for utilizing AdvAug. It seems that injecting adversarial perturbations into the ``high-level'' feature embeddings in general benefits more to mitigate the overfitting issue in data-limited regimes.
\end{itemize}
\vspace{-2mm}
In summary, we observe that applying AdvAug to the last layer of $\mathcal{D}$ and the first layer of $\mathcal{G}$, with PGD-1 and step size 0.01, seems to be a sweet-point configuration for data-efficient GAN training, which is hence adopted as our default setting.

\begin{wraptable}{r}{0.45\linewidth}
\vspace{-1.8em}
\caption{\small Performance of SNGAN models augmented by Gaussian Noise or AdvAug on $100\%$ and $10\%$ training data.}
\label{tab:random_noise}
\centering
\resizebox{1\linewidth}{!}{
\begin{tabular}{l|cc|cc}
\toprule
\multirow{2}{*}{Methods} & \multicolumn{2}{c}{100\% training data} & \multicolumn{2}{|c}{10\% training data} \\ \cmidrule(lr){2-3} \cmidrule(lr){4-5}
& IS & FID & IS & FID \\ \midrule
Baseline & 8.29 & 15.69 & 5.24 & 44.22\\
+ Gaussian Noise & 8.30 & 14.52 & 5.53 & 44.86\\
+ AdvAug & 8.42 & 13.99 & 6.10 & 41.25\\
\bottomrule
\end{tabular}}
\vspace{-4mm}
\end{wraptable}

\vspace{-0.1em}
\paragraph{Comparison with Baselines.} Naive baselines, i.e., random pruning, one-shot magnitude pruning (OMP) \cite{han2015deep,chen2021gans}, network slimming (NS \cite{liu2017learning}, and random noise augmentation, are evaluated in Figure~\ref{fig:aba} (\textit{Right}) and Table~\ref{tab:random_noise}. Compared to random pruning, OMP and NS, IMP produces much better GAN tickets, especially at high sparsity levels (e.g., $\ge 48.8\%$). Compared to augmenting features with random noise sampled from $\mathcal{N}(0,0.01^2)$, AdvAug also achieves larger performance gains on both $100\%$ and $10\%$ training data regimes. 

\vspace{-0.1em}
\section{Conclusion and Discussion of Broader Impact} \label{sec:conclusion}
% \vspace{-0.5em}
We introduce a novel perspective for data-efficient GAN training by leveraging lottery tickets, which augmentations can further enhance, including our newly introduced feature-level augmentation. Comprehensive experiments consistently demonstrate the effectiveness of our proposal, on diverse GAN architectures, objectives, and datasets. 
Note that although finding lottery tickets requires a costly train-prune-retrain process, only \emph{data efficiency} is of interest in this work. An intriguing future work would be to pursue data and resource efficiency (training and inference) together.

This research aims to enhance GAN training in the limited data regimes. However, it might amplify the existing societal risk of applying GANs. For example, the issue of image generation bias may be impacted or even amplified by the sparse structures, which we will verify in future work. The data-efficient generation ability might also be leveraged by undesired applications such as DeepFake.

% \clearpage

\bibliographystyle{unsrt}
\bibliography{ref}

\clearpage

\appendix

\renewcommand{\thepage}{A\arabic{page}}  
\renewcommand{\thesection}{A\arabic{section}}   
\renewcommand{\thetable}{A\arabic{table}}   
\renewcommand{\thefigure}{A\arabic{figure}}

%%%%%%%%%%%%%%%%%%%%%%%%%%%%%%%%%%%%%%%%%%%%%%%%%%%%%%%%%%%%
% \clearpage
\section*{Checklist}

\begin{enumerate}

\item For all authors...
\begin{enumerate}
  \item Do the main claims made in the abstract and introduction accurately reflect the paper's contributions and scope?
    \answerYes{}
  \item Did you describe the limitations of your work?
    \answerYes{Please see Section~\ref{sec:conclusion}.}
  \item Did you discuss any potential negative societal impacts of your work?
    \answerYes{Please see Section~\ref{sec:conclusion}.}
  \item Have you read the ethics review guidelines and ensured that your paper conforms to them?
    \answerYes{}
\end{enumerate}

\item If you are including theoretical results...
\begin{enumerate}
  \item Did you state the full set of assumptions of all theoretical results?
    \answerNA{Our work does not have theoretical results.}
	\item Did you include complete proofs of all theoretical results?
    \answerNA{Our work does not have theoretical results.}
\end{enumerate}

\item If you ran experiments...
\begin{enumerate}
  \item Did you include the code, data, and instructions needed to reproduce the main experimental results (either in the supplemental material or as a URL)?
    \answerYes{All used datasets are publicly available, and we follows the standard instructions in the cited papers, as shown in Section~\ref{sec:exp_details} and~\ref{sec:more_implement_details}. All of our codes are included in \url{https://github.com/VITA-Group/Ultra-Data-Efficient-GAN-Training}.}
  \item Did you specify all the training details (e.g., data splits, hyperparameters, how they were chosen)?
    \answerYes{All training details are provides in Section~\ref{sec:exp_details}.}
	\item Did you report error bars (e.g., with respect to the random seed after running experiments multiple times)?
    \answerYes{There independent evaluations are conducted. Meanwhile, the average performance with their standard deviations are reported in our paper.}
	\item Did you include the total amount of compute and the type of resources used (e.g., type of GPUs, internal cluster, or cloud provider)?
    \answerYes{The description of adopted computing resources are collected in Section~\ref{sec:exp_details}.}
\end{enumerate}

\item If you are using existing assets (e.g., code, data, models) or curating/releasing new assets...
\begin{enumerate}
  \item If your work uses existing assets, did you cite the creators?
    \answerYes{We use the public datasets and also cite their creators, as shown in Section~\ref{sec:exp_details} and~\ref{sec:more_implement_details}.}
  \item Did you mention the license of the assets?
    \answerNo{The licenses of the datasets are provided in the cited papers.}
  \item Did you include any new assets either in the supplemental material or as a URL?
    \answerYes{All used datasets are public available. All of our codes are included in \url{https://github.com/VITA-Group/Ultra-Data-Efficient-GAN-Training}.}
  \item Did you discuss whether and how consent was obtained from people whose data you're using/curating?
    \answerNA{We did not use/curate new data.}
  \item Did you discuss whether the data you are using/curating contains personally identifiable information or offensive content?
    \answerNA{We only use public and widely adopted datasets in this paper. We do not think there are any issues of personally identifiable information or offensive content.}
\end{enumerate}

\item If you used crowdsourcing or conducted research with human subjects...
\begin{enumerate}
  \item Did you include the full text of instructions given to participants and screenshots, if applicable?
    \answerNA{}
  \item Did you describe any potential participant risks, with links to Institutional Review Board (IRB) approvals, if applicable?
    \answerNA{}
  \item Did you include the estimated hourly wage paid to participants and the total amount spent on participant compensation?
    \answerNA{}
\end{enumerate}

\end{enumerate}

%%%%%%%%%%%%%%%%%%%%%%%%%%%%%%%%%%%%%%%%%%%%%%%%%%%%%%%%%%%%

\section{More Methodology Details}\label{sec:more_methods}

\subsection{More about Feature-level Augmentation in GANs via Adversarial Training}

\paragraph{Adversarial feature-level augmentation on generator $\mathcal{G}$.} Denote the generator as $\mathcal{G}=\mathcal{G}_2\circ\mathcal{G}_1$. Adversarial perturbations $\hat\delta$ generated by PGD, are applied to the intermediate feature space $\mathcal{G}_1(\boldsymbol{z})$, which can be depicted as follows:  
\begin{align}
&\mathcal{L}^{\mathrm{adv}}_{\mathcal{G}}\Def\max_{\|\hat\delta\|_{\infty}\le\epsilon}\mathbb{E}_{\boldsymbol{z}\sim p(\boldsymbol{z})}[f_{\mathcal{G}}(-\mathcal{D}(\mathcal{G}_2(\mathcal{G}_1(\boldsymbol{z})+\hat\delta)))], \label{eq:adv_g} \\ 
&\min_{\theta}\mathcal{L}_\mathcal{G}+\lambda_1\cdot\mathcal{L}^{\mathrm{adv}}_{\mathcal{G}}, \label{eq:min_adv_g}
\end{align}
where $\lambda_1$ controls the influence of adversarial information. We choose $\lambda_1=1$ tuned by a grid search.

\vspace{-2mm}
\paragraph{Adversarial feature-level augmentation on discriminator $\mathcal{D}$.} Denote the discriminator as $\mathcal{D}=\mathcal{D}_2\circ\mathcal{D}_1$. We augment  features of both real and generated samples. Specifically,
\begin{align}
\mathcal{L}^{\mathrm{adv}}_{\mathcal{D}}\Def & \min_{\|\delta\|_{\infty}\le\epsilon}\mathbb{E}_{\boldsymbol{x}\sim p_{\mathrm{data}}(\boldsymbol{x})}[f_{\mathcal{D}}(-\mathcal{D}_2(\mathcal{D}_1(\boldsymbol{x})+\delta))] + \nonumber \\
& \min_{\|\hat\delta\|_{\infty}\le\epsilon}\mathbb{E}_{\boldsymbol{z}\sim p(\boldsymbol{z})}[f_{\mathcal{D}}(\mathcal{D}_2(\mathcal{D}_1(\mathcal{G}(\boldsymbol{z}))+\hat\delta))], \label{eq:adv_d} \\
& \hspace{-12mm} \max_{\phi}\mathcal{L}_\mathcal{D}+\lambda_2\cdot\mathcal{L}^{\mathrm{adv}}_{\mathcal{D}}, \label{eq:max_adv_d}
\end{align}
where adversarial perturbations $\delta$ and $\hat\delta$ are applied to intermediate features $\mathcal{D}_1(x)$ and $\mathcal{D}_1(\mathcal{G}(z))$, respectively. $\lambda_2$ balances the effects of clean features and adversarial augmented features. In our case, $\lambda_2=1$ tuned by a grid search.

\paragraph{The overall pipeline of AdvAug.} As presented in Figure~\ref{fig:AdvAug_gan}, we augment the intermediate features of both the discriminator and generator. First, for augmenting $\mathcal{D}$, it minimizes $\mathcal{L}_{\mathcal{D}}^{\mathrm{adv}}$ to craft the adversarial perturbations for features from both real data and generated samples, and then maximizes $\mathcal{L}_{\mathcal{D}}$ together with $\mathcal{L}_{\mathcal{D}}^{\mathrm{adv}}$ to update the discriminator according to Eqn.~\ref{eq:max_adv_d}. Augmenting $\mathcal{G}$ works similarly, but only on generated samples' features $\mathcal{G}_1(\boldsymbol{z})$. The full algorithm of training GAN with both \textbf{data- and feature-level augmentations} is summarized in Algorithm~\ref{alg:AdvAug}.

\section{More Implementation Details} \label{sec:more_implement_details}

\subsection{More Details of Adopted Datasets}
\paragraph{Complete descriptions.} The CIFAR-10 and CIFAR-100 datasets each consist of $60,000$ $32\times32$ color images in $10/100$ classes, with $6,000/600$ images per class, respectively. The ratio between the number of training and testing images is $5:1$. Tiny-ImageNet contains $200$ image classes, a training/validation/test dataset of $100,000$/$10,000$/$10,000$ $64\times64$ images. ImageNet has $1,000$ image classes, $1,281,167$ training samples, and $50,000$ validation samples. In all experiments, we use $128\times128$ resolution for ImageNet samples.

\paragraph{Download links.} We list the download links for adopted datasets as follows:

($i$)~CIFAR-10/100:~\url{https://www.cs.toronto.edu/~kriz/cifar.html}

($ii$)~Tiny-ImageNet:~\url{https://www.kaggle.com/c/tiny-imagenet}

($iii$)~ImageNet:~\url{http://www.image-net.org}

($iv$)~Few-shot datasets \cite{zhao2020diffaugment}: \url{https://hanlab.mit.edu/projects/data-efficient-gans/datasets/}

\paragraph{Train-val-test splitting and subset constructions.} We follow the official splitting in the datasets. To construct subsets for the limited-data GAN training, we randomly sample a certain portion (e.g., $10\%$) from full training sets.

\subsection{More Details of Reported Sparsity}

\paragraph{How is the sparsity level selected?} We perform iterative magnitude pruning which each time removes a fixed portion (e.g., $20\%$) of the remaining weights with the smallest magnitudes, leading to the series of sparsity levels like \{$20\%$ ($1-1\times0.8$), $36\%$ ($1-1\times0.8^2$), $49\%$ ($1-1\times0.8^3$), $59\%$ ($1-1\times0.8^4$), $67\%$ ($1-1\times0.8^5$)\}. It is a widely adopted fashion in the literature~\citep{Frankle:2019vz,chen2021gans} of the lottery tickets hypothesis, and we strictly follow the standard convention.   

\section{More Experimental Results} \label{sec:more_res}

\paragraph{Comparisons with a smaller network baseline.} To show our achieved improvements not only come from the reduced network capacity but also from the sparse topology, we implement the ``small-dense” baseline by shrinking the number of channels and constraining its number of parameters to be equivalent to that of the sparse subnetwork. We take $67.24\%$ sparse BigGAN on $10\%$ training data of CIFAR-100 as the experimental setup. Then we train them together with DiffAug and AdvAug, and report the (FID$\downarrow$). LTH : Random Pruning : small-dense : Dense = $22.37$ : $25.73$ : $23.58$ : $23.94$. The results indicate the small-dense baseline with reduced sample complexity is helpful ($23.58$ v.s. $23.94$), while most of the benefits come from the identified sparse structure of winning tickets ($22.37$ v.s. $23.94$). The sparse structure of subnetworks matters. Meanwhile, we notice that recent literature also share consistent findings: $(i)$ big models are better few-shot learners~\cite{chen2020big}; $(ii)$ big models produce better winning lottery tickets~\cite{chen2020lottery2}.

\paragraph{Generalization study of our proposal.} Our framework is generalizable across diverse GAN architectures, which is also carefully evidenced in our main text (i.e., SNGAN, BigGAN, StyleGAN-v2). To further demonstrate it, we conduct extra experiments to combine our training framework with the proposed GAN architecture (i.e., + skip + decode) from~\cite{liu2020towards}. We observe that sparse GAN tickets at $36\%$ sparsity with augmentations further obtain ($2.03$,$0.75$,$0.26$) FID reductions on (Obama, Grumpy cat, Panda), which again validates the effectiveness of our proposal.

\paragraph{Pruning and augmentations on baseline pre-trained methods.} We apply our proposed training framework (LTH pruning + augmentations) to the baseline pre-trained method in Table~\ref{tab:fewshot}. The performance of FreezeD with a pre-trained StypleGAN-V2 is collected in Table~\ref{tab:more_fewshot}. We find consistent observations that our training framework (LTH pruning + augmentations) benefits FreezeD on few-shot generation tasks.

\begin{table}[t]
\caption{\small FreezeD~\cite{mo2020freeze} results with/without our proposed training framework.}
\vspace{-2mm}
\label{tab:more_fewshot}
\centering
\small
%\resizebox{1\textwidth}{!}{
\begin{tabular}{lccccc} 
\toprule
\multirow{2}{*}{Methods} & \multicolumn{3}{c}{100-shot by \cite{zhao2020diffaugment}} & \multicolumn{2}{c}{AnimalFace} \\ \cmidrule(lr){2-4} \cmidrule(lr){5-6}
& Obama & Grumpy Cat & Panda & Cat & Dog \\ \midrule
FreezeD ($0.00\%$) & 41.87 & 31.22 & 17.95 & 47.70 & 70.46\\ 
FreezeD ($0.00\%$) + DiffAug + AdvAug & 36.52 & 30.04 & 16.23 & 46.39 & 64.21\\ 
FreezeD ($48.80\%$) & 40.10 & 30.16 & 16.52 & 46.58 & 66.74\\ 
FreezeD ($48.80\%$) + DiffAug + AdvAug & 35.25 & 29.62 & 15.19 & 45.94 & 61.30\\ 
\bottomrule
\end{tabular}
%}
\vspace{-1.5mm}
\end{table}

\begin{table}[t]
\caption{\small Transfer performance of winning tickets found with FreezeD~\cite{mo2020freeze} and StyleGAN-V2 on FFHQ.}
\vspace{-2mm}
\label{tab:transfer}
\centering
\small
%\resizebox{1\textwidth}{!}{
\begin{tabular}{lccccc} 
\toprule
\multirow{2}{*}{Methods} & \multicolumn{3}{c}{100-shot by \cite{zhao2020diffaugment}} & \multicolumn{2}{c}{AnimalFace} \\ \cmidrule(lr){2-4} \cmidrule(lr){5-6}
& Obama & Grumpy Cat & Panda & Cat & Dog \\ \midrule
StyleGAN-V2 finetune ($0.00\%$) & 41.87 & 31.22 & 17.95 & 47.70 & 70.46\\ 
StyleGAN-V2 finetune ($0.00\%$) + DiffAug + AdvAug & 36.52 & 30.04 & 16.23 & 46.39 & 64.21\\ 
StyleGAN-V2 finetune ($48.80\%$) & 41.33 & 30.68 & 16.47 & 46.75 & 68.50\\ 
StyleGAN-V2 finetune ($48.80\%$) + DiffAug + AdvAug & 35.90 & 29.73 & 14.86 & 46.01 & 63.15 \\ 
\bottomrule
\end{tabular}
%}
\vspace{-1.5mm}
\end{table}

\begin{table}[t]
\caption{\small FID ($\downarrow$) and IS ($\uparrow$) results of SNGAN with $10\%$ training data of CIFAR-10 at diverse sparsity levels. The setting ``Sparse Tickets + Aug" is reported here.}
\vspace{-2mm}
\label{tab:more_sparsity}
\centering
\small
\resizebox{1\textwidth}{!}{
\begin{tabular}{lccccccccccc} 
\toprule
Sparsity & Dense ($0\%$) & $5\%$ & $10\%$ & $15\%$ & $20\%$ & $25\%$& $30\%$ & $35\%$ & $40\%$ & $45\%$ & $50\%$\\ \midrule
FID ($\downarrow$) & 41.25 & 39.31 (↓1.94) & 36.85 (↓4.40) & 34.09 (↓7.16) & 32.47 (↓8.78) & 33.62 (↓7.63) & 36.38 (↓4.87) & 35.16 (↓6.09) & 35.94 (↓5.31) & 36.40 (↓4.85) & 37.22 (↓4.03) \\ 
IS ($\uparrow$) & 5.64 & 5.72 (↑0.08) & 5.87 (↑0.23) & 6.03 (↑0.39) & 6.20 (↑0.56) & 6.14 (↑0.50) & 5.97 (↑0.33) & 5.93 (↑0.29) & 5.96 (↑0.32) & 5.91 (↑0.27) & 6.01 (↑0.37) \\ 
\bottomrule
\end{tabular}
}
\vspace{-1.5mm}
\end{table}

\paragraph{Mask transferring.} As demonstrated in~\cite{chen2020lottery,chen2020lottery2}, the winning tickets found on the pre-training task, show impressive transferability to diverse downstream tasks. We conduct similar pre-training and transfer studies in our context. Precisely, we first identify a ``pre-training” GAN winning ticket with the FreezeD method~\cite{zhao2020diffaugment} and the StyleGAN-V2 backbone on the FFHQ dataset. Then, we fine-tune it on diverse few-shot domains and report their performance in Table~\ref{tab:transfer}. We find that in this practical and meaningful pre-training + fine-tuning scheme, our proposed LTH pruning + augmentations method is still effective.

\paragraph{Fine-grained sparsity levels.} To demonstrate our proposal's effectiveness across diverse sparsity levels, we adjust the pruning ratios so that each time we remove $5\%$ of the total weights with the smallest magnitudes, and conduct extra experiments on these sparsity levels \{$5\%$, $10\%$, $15\%$, $20\%$, $25\%$, $30\%$, $35\%$, $40\%$, $45\%$, $50\%$\}. Results in Table~\ref{tab:more_sparsity}, evidence the consistent benefits from our proposed training pipeline. All experimental configurations are the same as the ones in Figure~\ref{fig:SNGAN}.

\end{document}